\documentclass{article}

     \usepackage[nonatbib]{neurips_2020}

\usepackage[utf8]{inputenc} 
\usepackage[T1]{fontenc}    
\usepackage{hyperref}       
\usepackage{url}            
\usepackage{booktabs}       
\usepackage{amsfonts}       
\usepackage{nicefrac}       
\usepackage{microtype}      
\usepackage{graphics,graphicx,epsfig,algorithm,algpseudocode}
\usepackage{latexsym,amsmath,amsfonts,amssymb,color}
\usepackage{graphicx}
\usepackage{physics}
\usepackage{threeparttable}
\usepackage{caption}
\usepackage{array}
\newcommand\tabcaption{\def\@captype{table}\caption}
\newcommand{\be}{\begin{eqnarray*}}
\newcommand{\ee}{\end{eqnarray*}}
\newcommand{\ibe}{\begin{eqnarray}}
\newcommand{\iee}{\end{eqnarray}}

\title{Disturbing Target Values for Neural Network Regularization}

\author{%
  Yongho Kim \\
  Institute of Computer Science\\ University of Hildesheim\\ Samelsonplatz 1, 31141 Hildesheim, Germany\\
  
}

\begin{document}

\maketitle
\begin{abstract}
Given the increasing computational capabilities of modern computing systems, we are seeing bigger and more complex neural networks being used. However with this increase the models have shown to perform well on training data but suffering on unseen test data, this problem is known as overfitting. This phenomenon is well recognized in recent researches that the model tends to become overparameterized when the networks get more complex. Diverse regularization techniques have been developed such as L2 regularization, Dropout, DisturbLabel (DL) to prevent overfitting. DL, a newcomer on the scene, regularizes the loss layer by flipping a small share of the target labels at random and training the neural network on this distorted data so as to not learn the training data. It is observed that high confidence labels during training cause the overfitting problem and DL selects disturb labels at random regardless of the confidence of labels. To solve this shortcoming of DL, we propose Directional DisturbLabel (DDL) a novel regularization technique that makes use of the class probabilities to infer the confident labels and using these labels to regularize the model. This active regularization makes use of the model behavior during training to regularize it in a more directed manner. To address regression problems, we also propose DisturbValue (DV), and DisturbError (DE). DE uses only predefined confident labels to disturb target values. DV injects noise into a portion of target values at random similar to DL. In this paper, 6 and 8 datasets are used to validate the robustness of our methods in classification and regression tasks respectively. Finally, we demonstrate that our methods are either comparable to or outperform DisturbLabel, L2 regularization, and Dropout. Also, we achieve the best performance in more than half the datasets by combining our methods with either L2 regularization or Dropout.
\end{abstract}

\section{Introduction}
Training computers to learn and think like human to produce reliable results has been a topical study in Machine Learning (ML). Though there are several types of ML, for the sake of simplicity, we will focus only supervised learning in which we train the model with known dataset to make predictions and let the trained model predict new unseen data. Two different tasks under supervised learning are separated as classification and regression. They share the same concept of mapping inputs to outputs, but differ in the form of targets. For classification, the output is discrete or categorical, while the output is continuous or real numbers for regression. Attempting to create machine to reason like human, the mathematical models mimicking the human brain was introduced and has become commonly known as Neuron Network (NN). The idea of NN was first introduced in 1948 by Alan Turing \cite{copeland04turing} and has been developed further forming a series of algorithm having a neuron-like network. These networks are used to solve similar, but more complex, problems to other ML algorithms such as credit card fraud detection in financial sector \cite{CHOUIEKH2018133} and an automated driving system in automotive industry \cite{9307303}. Artificial Neural Network (ANN), Recurrent Neural Network (RNN) or Convolutional Neural Network are some examples of these neuron-like algorithms. Deeper networks yield higher accuracy and higher ability to data representation. However, it comes with large numbers of parameters which makes the models prone to overfitting or fail to generalize unseen data. This phenomenon is likely to happen when the model is very complex while training data is insufficient.  Generally, deeper networks are deployed and they have far more parameters than LeNet  \cite{NIPS1989_53c3bce6} does. Overfitting phenomenon conflicts with objective of any ML algorithm where we need the trained model to perform well not only on training data, but also on the unseen data. To avoid such event, regularization is brought into play. 

In recent years, several regularization techniques have been developed to be applied to various parts of the network. Some techniques are applied to weight such as \textit{DropConnect} \cite{pmlr-v28-wan13} that drops the weight between connected nodes leaving the connected layer sparse but the node can still be active. \textit{$L_2$-regularization or weight decay} \cite{Krizhevsky2009LearningML} , adds penalty term with a hyperparameter ($\lambda$) to the error function resulting in weight decaying close to zero. \textit{Data augmentation} \cite{krizhevsky2012imagenet} is used at the input layer to perform image transformation such as flipping, zooming, shifting and cropping. \textit{Dropout} \cite{srivastava2014dropout} can be implemented, for instance, at the hidden layers to reduce the dependence between neurons by randomly dropping out the nodes from the network. \textit{Penalizing confident output} by  \cite{pereyra2017regularizing} is done at the output layer which introduces negative entropy to the negative likelihood during training.  Last example is from the latest novel idea concerns the loss layer. \textit{DisturbLabel(DL)} \cite{xie2016disturblabel} aims to attack the loss layer by randomly flipping the ground truth with a shared hyperparameter ($\alpha$). Among the numbers of various regularization methods, DL is claimed as the novel algorithm attacking the loss layer for classification task. Its simplicity of implementation, yet proving the compelling result comparable to the widely known technique such as dropout, has convinced us to explore and generate new prospects inherited from this concept. In this paper, we improve the regularization technique on the loss layer for classification task and extend the idea onto regression task. 

For classification, we propose \textbf{Directional Disturb Label (DDL)} which is the method of selecting systematically which labels to disturb by excluding non-confident labels from the candidates based on cosine similarity. We show that this improvement can reduce misclassification rate comparing to the baseline and performs better with deeper network. Additionally, the experimental results reaffirm that there is no burden on the cooperation of DDL with other regularization methods. For regression, we propose \textbf{DisturbValue(DV)} and \textbf{DisturbError(DE)}, the two novel methods developed from applying disturbing procedure of DL onto regression task. The experimental results show the efficiency of our methods. Our codes are available from the first author's Github (https://github.com/kimy-de/DisturbMethods).

\noindent Our main contributions include:
\begin{enumerate}
    \item {Propose model-agnostic noisy regularization methods.}
    \item{Improve DisturbLabel by filtering non-confident labels from the candidates of disturb labels.}
    \item{Demonstrate the robustness of our methods using 14 datasets.}
\end{enumerate}

\section{Related Work}\label{related_work}

Regularization methods come in a wide variety of the areas of the neural network they are aimed at regularization can be imposed on weights, hidden or inputs nodes of the neural network, but regularizing within the loss layer is relatively new. Xie, Lingxi et al.\cite{xie2016disturblabel} introduced DisturbLabel (DL) which is the first work investigating this area for convolutional neural networks. 

The closest method to DisturbLabel is \textit{label smoothing} \cite{szegedy2015rethinking}, which also perturbs the ground-truth labels by softening it to a vector of probabilities of belonging to each of the classes in the task, while DisturbLabel flips the label fully. As the authors point out, the main advantage of DisturbLabel is being stochastic while soft labelling is deterministic and therefore cannot provide the same strong regularization as by DisturbLabel.

Taejong Joo et al.\cite{joo2020being} also perturbs ground-truth labels similarly to label smoothing, but use Bayesian approach, and instead of manipulating the labels directly, they consider the ground truth to be a random variable of a categorical probability over class labels rather than being given by the training label.

From recent findings on manipulating the labels, the \textit{mixup} method \cite{zhang2017mixup} is one of the prominent. It suggests training a neural network on convex combinations of pairs of examples and their labels.
Weizhi Li et al.\cite{li2020regularization}  combine  the label smoothing with the hypothesis that more confident predictions require stronger regularization (we also exploit this hypothesis in our work). They perform \textit{regularization via structural label smoothing}, imposing various smoothing strength on clusters of data lying in different parts of feature space.

Noisy regularization for regression tasks are well known techniques, but as pointed out in a survey paper\cite{moradi2020survey}, the noise is added either to inputs (\cite{poole2014analyzing}) or to weights (\cite{hochreiter1995simplifying}). To the best of our findings, adding noise to the targets as a form of regularization has been never attempted before. Ehsan Imani et al.\cite{imani2018improving} did add Gaussian noise to targets, but their experiments are different from our proposed DV approach in the sense that (a) it was done as a form of augmentation and (b) noise was added to all targets.

Adding Gaussian noise to target values was also explored \cite{wang1999training}, but our works compliment rather than repeat each other. First, this prior work concentrates on the convergence properties of the network with noise added to the desired signal and does not investigate regularization effects. Second, they add noise to target values and use annealing schedule of the step size to control the variance of noise since the noise affects weight updating. On the contrary, we control the number of noisy values using noise rate instead of adjustment of the step size.

\section{Methodology}\label{method}
\subsection{Classification}
DisturbLabel (DL) \cite{xie2016disturblabel} regularizes the loss layer by replacing some ground truth labels with incorrect labels in each iteration. DL selects $k$ substitutes randomly to generate disturb labels based on a noise rate $\alpha\%$ that determines the amount of $k=\frac{\alpha}{100}\times(batch\, size)$. When $\alpha=0\%$, a batch-training works without any change. When $\alpha>0\%$, each of selected true labels $y$ is converted into a disturb label by a Multinoulli distribution with the following probabilities:
\begin{equation} 
p_c=1-\frac{(C-1)\alpha}{100C},\, p_i=\frac{\alpha}{100C}
\end{equation}
where $p_c$ and $p_i$ replaces 1 and 0 respectively in the one-hot vector and C is the number of classes. The DL does not consider the confidence of labels to generate disturb labels. 

\begin{figure}[t!]
\centering
\includegraphics[width=0.7\columnwidth]{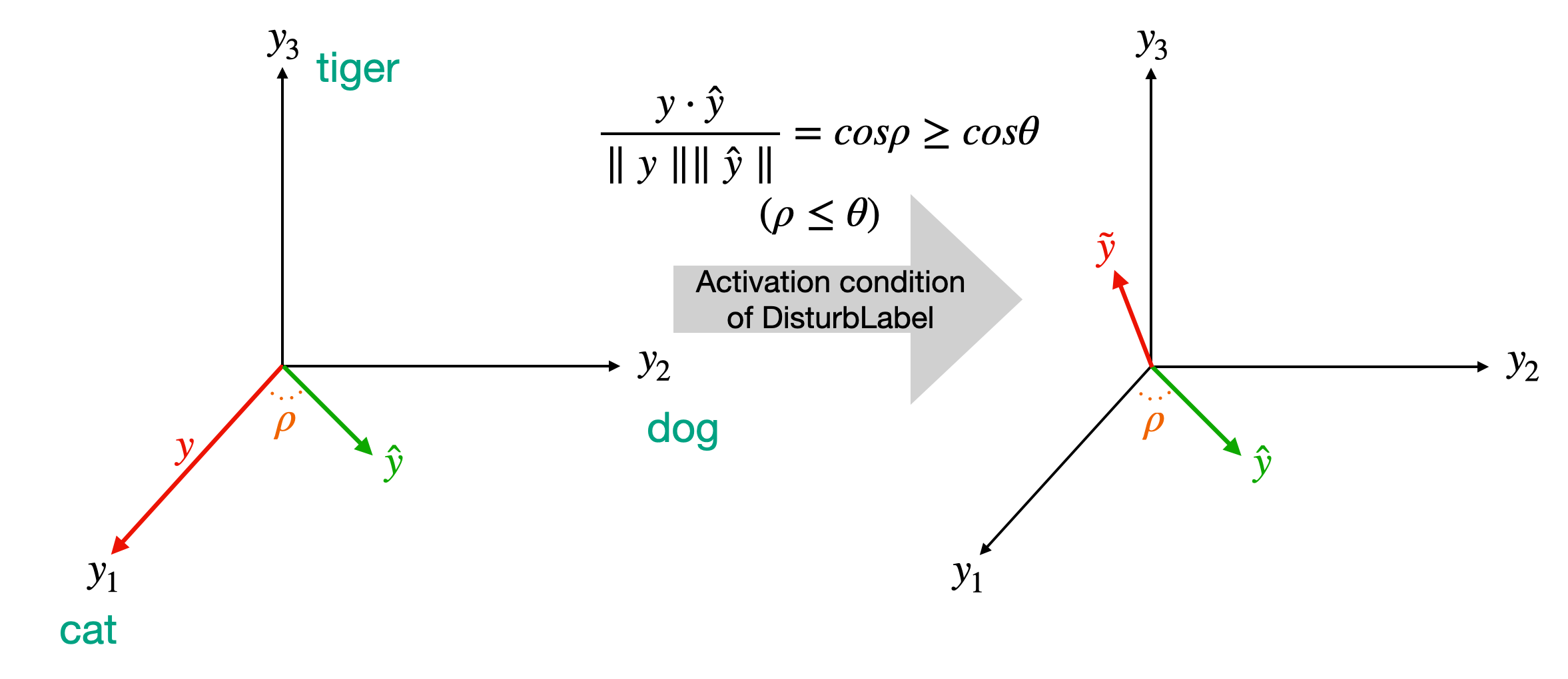}
\caption{Schematic representation of prediction vectors}%
\label{ddl}
\end{figure}

Our hypothesis is that high confident labels cause the overfitting problem, hence our method \textbf{Directional DisturbLabel (DDL)} considers only confident labels as candidates to be disturbed. To be specific, we define the confident label as $\hat{y}$ satisfying $\frac{y\cdot\hat{y}}{\parallel y \parallel\parallel \hat{y}\parallel}\geq cos\theta$ where $\theta$ is a permissible angle. In $\mathbb{R}^C$, the angle $\rho$ between $y$ and $\hat{y}$ is calculated by the cosine similarity and we say that $\hat{y}$ is a confident label if $\rho <\theta$. To simplify the formula $\frac{y\cdot\hat{y}}{\parallel y \parallel\parallel \hat{y}\parallel}\geq cos\theta$, we use the unit vectors, $u=\frac{\hat{y}}{\parallel \hat{y}\parallel}$ and $y$, with $\theta=\frac{\pi}{3}$ so that $y\cdot u\geq 0.5$ is applied to select substitutes for disturb labels in each batch data. The range of $\rho$ is $[0,\frac{\pi}{2}]$. $\rho=0$ when $\hat{y}$ is the same as $y$(i.e. the same direction). Plus, $y$ and $\hat{y}$ always have positive elements so that the maximum angle between them is $\frac{\pi}{2}$. In our setting,  non-confident labels from the candidates are excluded by the non-confident interval of $(\frac{\pi}{3},\frac{\pi}{2}]$. For instance, Figure \ref{ddl} shows that all the outputs of classification models with three classes are on the three dimensional space and the natural basis of the vector space are the one-hot vectors of the three true labels, $(1,0,0),(0,1,0)$, and $(0,0,1)$. Therefore, there exist $y$ and $\hat{y}$ are on the same space such that $y\cdot\hat{y}\geq 0.5$ is calculated.

\subsection{Regression}
\noindent Our hypothesis is that noise injection to target values regularizes the classifier layer by the oscillation of target values because the measurement of continuous target values could have errors caused by machine tolerance, human ability, measurement condition, and so on. For example, when we measure the current temperature, we can think that $16.55^\circ$C and $16.54^\circ$C are the same depending on the tolerance of thermometers so proper noise injection to target values can lead to a regularization effect without changing the attribute of target values. Thus, we use the concept of DisturbLabel (DL) for regression tasks. However, the original DL can be applied in classification problems so we propose two disturb methods by extending the concept of DL to regression problems. Given a mini-batch set $D_i \subset D=\{ (\vb{x_n}, y_n) \mid \vb{x_n}\in\mathbb{R}^p, y_n\in\mathbb{R}, n, p\in\mathbb{N}  \}$. Then a prediction $\hat{y}=f(\vb{x};\theta)$ where $f$ is a regression model with model parameters $\theta$.

\begin{figure}[t!]
\centering
\includegraphics[width=1\columnwidth]{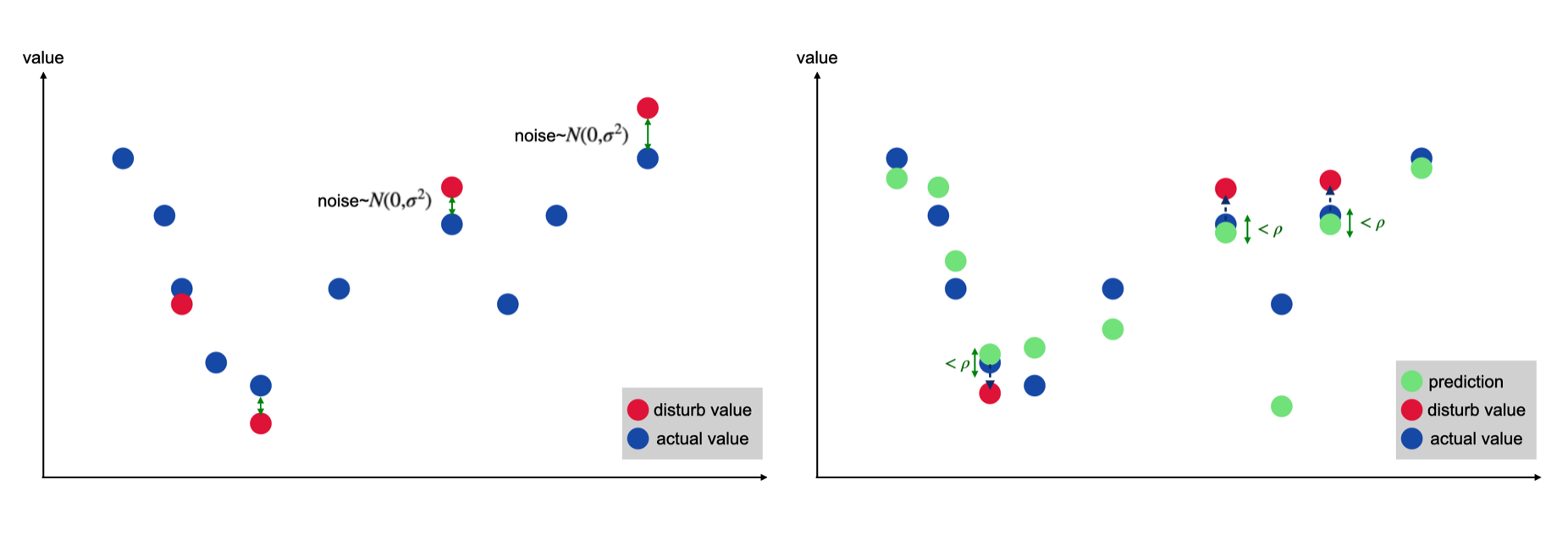}
\caption{Illustration of DisturbValue(left) and DisturbError(right)}%
\label{fig:de}
\end{figure}

To prevent overfitting in the training session, \textbf{DisturbValue (DV)} adds Gaussian noise to some of the target values at random. To be specific, a target value $y$ is replaced by $\tilde{y}=y+\epsilon$ following $\epsilon\sim N(0, \sigma^2)$ based on a noise rate $\alpha$. When $\alpha=p\%$, $\epsilon$ is added randomly to $k$ target values in each batch data where $k \leq \frac{p}{100} \times  (batch\, size)$.
As hyperparameters of DV, $\alpha$ and $\sigma$ have a huge impact on the regularization performance, but it is too hassle for controlling them depending on datasets. To reduce the number of the hyperparameters, we transform the domain of target values into $[0,1]$ by the MinMax scaler $f(x)=\frac{x-min(x)}{max(x)-min(x)}$ and $\sigma=0.01$ is fixed as a default. Thus, all of the domain of target values is $[0,1]$ regardless of datasets so that the same standard variance $\sigma$ can be used for any datasets. Finally, we consider the only one hyperparameter $\alpha$ during the training.

DV adds noise to target values at random. However, \textbf{DisturbError (DE)} adds Gaussian noise $\epsilon\sim N(0, \sigma^2)$ to target values satisfy $\mid y- \hat{y}\mid < \rho$ where $ \rho$ is a residual boundary. We say that $\hat{y}$ is a high confident value if there exists small constant $\rho$ such that $\mid y- \hat{y}\mid < \rho$. When there are many high confident values, it is highly possible that a model is overfitted to train data. To prevent overfitting, DE disturbs the error of high confident values by redefining the error as $\mid y+\epsilon- \hat{y}\mid$. In DE, we should control $\rho$ and $\sigma$ originally, however we can consider only $\rho$ based on the same scaling and $\sigma$ as DV. 

Therefore, we find an optimal hyperparameter of DV and DE depending on datasets and compare to other regularization techniques in Section \ref{exp}. In this paper, mean square loss function $L=\frac{1}{N}\sum_{i=1}^{N}(\hat{y_i}-y_i)^2$ is used such that  $\tilde{L}=\frac{1}{N}\sum_{i=1}^{N}(\hat{y_i}-(y_i+\epsilon_i))^2$ is defined as an objective function with our disturb methods. Considering the gradient of $\tilde{L}$, \begin{equation} 
\frac{\partial\tilde{L}}{\partial\theta} = \frac{2}{N}\sum_{i=1}^{N}(\hat{y_i}-(y_i+\epsilon_i))\frac{\partial\hat{y_i}}{\partial\theta}\\
= \frac{2}{N}\sum_{i=1}^{N}((\hat{y_i}-y_i)-\epsilon_i)\frac{\partial\hat{y_i}}{\partial\theta}\\
= \frac{\partial L}{\partial\theta}- \frac{2}{N}\sum_{i=1}^{N}\epsilon_i\frac{\partial\hat{y_i}}{\partial\theta}
\end{equation}
is derived. It shows that the noise $\epsilon$ controls the gradient of prediction values to prevent overfitting. 

\section{Experiments}
\label{exp}
\subsection{Classification Task}

\begin{table}[tbp]
\caption{Description of the datasets for classification task} 
\label{clsDatasets}
\begin{threeparttable}
    \renewcommand{\arraystretch}{1.5}
    \fontsize{7pt}{7pt}\selectfont
\begin{tabular}{p{2cm}|p{1.5cm}p{1.5cm}p{1.5cm}p{1.5cm}p{1.5cm}p{1.5cm}}
	\toprule
	\textbf{Dataset} & \textbf{MNIST}\tnote{*}  & \textbf{FMNIST} & \textbf{CIFAR10}\tnote{*} & \textbf{CIFAR100} & \textbf{Intel} & \textbf{Art} \\
	\midrule
	
	\# classes & 10 & 10 & 10 & 100 & 6 & 5 \\
	\hline
		Size  of images & 28$\times$28& 28$\times$28 & 32$\times$32&  32$\times$32 & 150$\times$150 &  227$\times$227\tnote{**}  \\
	\hline
	\# channels & 1 & 1 & 3& 3 & 3 & 3 \\
	\hline
	\# instances & 70K & 70K & 60K  & 60K & 17K & 9K \\
	\hline
	Train/Test (\%
	)& 86/14  & 86/14 & 83/17 & 83/17 &  82/18&  86/14\\
	\bottomrule
\end{tabular} 
  \begin{tablenotes}[normal,flushleft] 
  \item[*] datasets also used by the baseline paper. CIFAR100 is implemented differently frrom the baseline. We use all 100 classes and treat the dataset as an additional one.
  \item[**] input size fed to the network (original sizes vary)
  \end{tablenotes}
\end{threeparttable}
\end{table}

\textbf{Evaluation metric.} We report test misclassification rate to compare the performance of the methods. Experiments for each dataset method combination were run for 5 times, average value alongside with standard deviation is reported.

\textbf{Baselines.} In our experiments we compare the performance of suggested DDL method against same baselines as the reference paper (including exploring cooperation of the methods) and DL method itself; \textit{no regularization, dropout, DistrubLabel (DL), DistrubLabel (DL) + dropout, Directional DisturbLabel (DDL), Directional DisturbLabel (DDL) + dropout.} 

\textbf{Datasets.}
Having a hypothesis that regularization effects should appear more vividly on 'easier' datasets, we conduct experiments on six collections of images of various complexity:
MNIST \cite{mnist2010}, FMNIST \cite{xiao2017/online}, CIFAR-10 \cite{CIFAR10}, CIFAR-100 \cite{Krizhevsky2009LearningML}, INTEL  \cite{intel} , and ART \cite{art}. The characteristics of the datasets are summarized in Table \ref{clsDatasets}.

\textbf{Architecture.} \textbf{\textit{LeNet}} \cite{NIPS1989_53c3bce6} is modified in accordance to baseline paper (two convolution units for MNIST and FMNIST dataset and three convolution units for CIFAR10, CIFAR100, ART and INTEL followed by ReLU and max pooling). \textbf{\textit{ResNet18}} \cite{resnet18} is modified with additional dropout after average pooling step for our experiment on models combination before feeding the input to the last layer with the softmax loss function.

\textbf{Optimizer.} The training procedure was performed using SGD optimizer with momentum and decaying learning rate. We start with learning rate of 0.001 and reduce it by factor 0.1 after 40, 60 and 80 epochs. Each experiment was run for 100 epochs in total.

\textbf{Hyperparameters.} This hyperparameter $\alpha$ (probability of the label to be disturbed) was tuned for each dataset separately. The optimal values of hyperparameters used are summarized in Table \ref{clsHP}. The dropout probability was set to 0.5 in all corresponding experiments.

\begin{table}[tbp]
	\caption{Optimal values of hyperparameter $\alpha$ (\%)} \label{clsHP}
\centering
\renewcommand{\arraystretch}{1.5}
	\fontsize{7pt}{7pt}\selectfont
\begin{tabular}{p{2cm}|p{1.2cm}|m{1.2cm}m{1.2cm}m{1.2cm}m{1.2cm}m{1.2cm}m{1.2cm}}
\toprule
	\textbf{Architecture} & \textbf{Method}	& \textbf{ MNIST }	&	\textbf{ FMNIST }	&	\textbf{ CIFAR10 }	&	\textbf{  CIFAR100  }	&	\textbf{  Intel  }	&	\textbf{   Art   }	\\
    \midrule
LeNet & DL	        &10		&	5	&	10	&	40	&	50	&	50	\\
& DDL	        &10		&	5	&	50	&	30	&	50	&	20	\\
\midrule
ResNet18 & DL	        &	20	&	5	&	10	&	10	&	5	&	10	\\
& DDL	        &	20	&	5	&	10	&	10	&	5	&	10	\\
\bottomrule
\end{tabular}
	\end{table}

\begin{table}[tbp]
	\caption[LeNet Experimental Result]{LeNet Experimental Result(average misclassification rate): DDL demonstrates good cooperation with Dropout and brings the best results for datasets most suffering from overfitting (less complex datasets). Results are reported as  \textbf{first} and  \underline{second} best.} \label{clsResultsLN}
\centering
	\renewcommand{\arraystretch}{1.5}
	\fontsize{7pt}{7pt}\selectfont
\begin{threeparttable}

\begin{tabular}{p{1.5cm}|p{1.6cm}p{1.6cm}p{1.6cm}p{1.6cm}p{1.6cm}p{1.6cm}p{1.6cm}}
    	\toprule
		& \textbf{MNIST} & \textbf{FMNIST} & \textbf{CIFAR10} & \textbf{CIFAR100} &\textbf{Intel} & \textbf{Art} \\
		\# classes&\textbf{10}&\textbf{10}&\textbf{10}&\textbf{100}&\textbf{6}&\textbf{5}\\
		\midrule
No reg.\tnote{1} 	&	0.86	$\pm$0.034	&	8.052	$\pm$0.112	&	24.82	$\pm$0.828	&	59.732	$\pm$1.590	&	15.614	$\pm$0.484	&	18.08	$\pm$0.597	\\
Dr.\tnote{2}	&	0.658	$\pm$0.049	&	7.784	$\pm$0.167	&	22.58	$\pm$0.771	&	\textbf{50.662	$\pm$0.828}	&	13.48	$\pm$0.575	&	16.372	$\pm$0.651	\\
DL	&	0.642	$\pm$0.052	&	\underline{7.748	$\pm$0.069}	&	23.77	$\pm$0.310	&	58.758	$\pm$0.596	&	13.289	$\pm$0.270	&	16.28	$\pm$0.335	\\
DDL	&	\underline{ 0.61	$\pm$0.041}	&	7.789	$\pm$0.111	&	22.614	$\pm$0.553	&	56.444	$\pm$0.337	&	13.542	$\pm$0.193	&	16.687	$\pm$0.948	\\
Dr.+DL	&	\textbf{0.58	$\pm$0.044}	&	7.816	$\pm$0.132	&	\textbf{21.65	$\pm$0.320}	&	50.87	$\pm$0.701	&	\underline{13.066	$\pm$0.339}	&	\underline{15.578	$\pm$0.456}	\\
Dr.+DDL	&	0.658	$\pm$0.086	&	\textbf{7.744	$\pm$0.210}	&	\underline{21.766	$\pm$0.403}	&	\underline{50.824	$\pm$1.487}	&	\textbf{12.366	$\pm$0.310}	&	\textbf{15.537	$\pm$0.498}	\\

		\bottomrule
	\end{tabular}
	\begin{tablenotes}
	\item[1] No reg. refers no to regularization
	\item[2] Dr. refers to dropout
	\end{tablenotes}
    \end{threeparttable}
\end{table}

\textbf{Experimental Results.} The results of the experimented on LeNet are summarized in Table \ref{clsResultsLN}. DDL method (in combination with Dropout) demonstrated the best results for half of the datasets and is in top-2 results for all of them. Intel and Art dataset considered as simpler datasets which have higher degree of overfitting the other more complex datasets. MNIST is also known for being easily classified by modern networks, therefore we can say that the results prove the hypothesis of DDL method having better regularization capacity and is useful on the datasets which needed it at most. The four newly tested datasets demonstrated similar behaviour except for CIFAR100, which is the most challenging of them, having far more classes. We assume that memorizing the ground truth was not really in place for CIFAR100, that is why perturbing the labels even more has not facilitated the training procedure. A more lightweight method - dropout - nevertheless has still improved the result compared to full absence of regularization.

The results of the experimented on ResNet18 is shown in Table \ref{clsResultsRN} confirming the results obtained for LeNet. For 5 out of 6 tested datasets, DDL (either alone or in combination with Dropout) has gained the best result compared to baselines and is second best for the remaining. FMNIST and CIFAR10 presumably do not require such strong regularization as the most simple MNIST, Intel and Art datasets, therefore using combination of DDL and dropout can be excessive in this case and DDL method obtains the best result alone. CIFAR100 still does not suffer overfitting even with the deeper network to the extend as the other datasets do and does not profit from strong regularization.

\begin{table}[tbp]
\caption[ResNet18 Experimental results]{ResNet18 Experimental results (average misclassification rate): With a more complex network, our method demonstrates vivid results against baselines. CIFAR100, however, does not require strong regularization and takes advantage of simpler DL. Results are reported as  \textbf{first} and  \underline{second} best.} \label{clsResultsRN}
    \begin{threeparttable}
    \centering
	\renewcommand{\arraystretch}{1.5}
	\fontsize{7pt}{7pt}\selectfont
    \begin{tabular}{p{1.5cm}|p{1.6cm}p{1.6cm}p{1.6cm}p{1.6cm}p{1.6cm}p{1.6cm}p{1.6cm}}
    		\toprule
		& \textbf{MNIST} & \textbf{FMNIST} & \textbf{CIFAR10} & \textbf{CIFAR100} &\textbf{Intel} & \textbf{Art} \\
		\# classes &\textbf{10}&\textbf{10}&\textbf{10}&\textbf{100}&\textbf{6}&\textbf{5}\\
		\midrule

No reg.\tnote{1}	&	0.668	$\pm$0.0618	&	8.354	$\pm$0.072	&	8.145	$\pm$0.430	&	30.557	$\pm$0.466	&   6.237	$\pm$0.269	&	2.927	$\pm$0.26	\\
Dr.\tnote{2}	&	0.601	$\pm$0.045	&	8.608	$\pm$0.179	&	\underline{7.813	$\pm$0.170}	&	28.363	$\pm$0.500	&	6.531	$\pm$0.333	&	\underline{2.837	$\pm$0.348}	\\
DL	&	0.612	$\pm$0.1567	&	\underline{8.244	$\pm$0.274}	&	7.99	$\pm$0.444	&	\textbf{28.048	$\pm$1.064}	&	\underline{6.004	$\pm$0.123}	&	2.909	$\pm$0.416	\\
DDL	&	0.558	$\pm$0.0507	&	\textbf{8.228	$\pm$0.107}	&	\textbf{7.667	$\pm$0.096}	&	28.236	$\pm$1.209	&	6.271	$\pm$0.207	&	2.891	$\pm$0.090	\\
Dr.+DL	&	\underline{0.543	$\pm$0.039}	&	8.480	$\pm$0.378	&	7.841	$\pm$0.198	&	28.123	$\pm$0.540	&	6.231	$\pm$0.346	&	3.035	$\pm$0.0342	\\
Dr.+DDL	&	\textbf{0.532	$\pm$0.054}	&	8.475	$\pm$0.191	&	8.111	$\pm$0.151	&	\underline{28.115	$\pm$1.186}	&	\textbf{5.917	$\pm$0.166}	&	\textbf{2.728	$\pm$0.206}	\\

		\bottomrule
	\end{tabular}

    \begin{tablenotes}
    \item[1]No reg. refers to no regularization, \item[2] Dr. refers to dropout.
    \end{tablenotes}
    \end{threeparttable}
    
	\end{table}
	
\newpage
\subsection{Regression Task}
\textbf{Evaluation metric.}  We use root-mean-square error (RMSE) \cite{HYNDMAN2006679}. For every experiment and every dataset we average RMSE for 20 runs and measure standard deviation.

\textbf{Baselines.} For all eight datasets we compare the results of our methods with the baselines which represent state-of-the-art on neural network regularization to the best of our knowledge and also combination of our methods with baselines: \textit{$L_2$ regularization, dropout, DV - Gaussian noise, DV - Laplacian noise, DV - cosine annealing \cite{inproceedings}, DE, DV + dropout, DV + $L_2$, DV + DE}.

\begin{table}[!t]
	\caption{Description of the datasets for regression task} \label{regdata}
\centering
\renewcommand{\arraystretch}{1.5}
	\fontsize{7pt}{7pt}\selectfont
\begin{tabular}{p{1.3cm}|m{1.2cm}m{1.2cm}m{1.2cm}m{1.2cm}m{1.2cm}m{1.2cm}m{1.2cm}m{1.2cm}}
\toprule
	\textbf{Dataset} & \textbf{BHP}	& \textbf{BS}	&	\textbf{AQ}	&	\textbf{MS}	&	\textbf{HP}	&	\textbf{SC}	&	\textbf{CC}	&	\textbf{AEP}\\
    \midrule
\# Instances	       		&	506	&	731	&	9,357	&	5,000	& 1460 & 21,263	&1,994 &19,735	\\
\# Feautres      		&	13	&	13	&	10	&	30	&	81	& 81 & 100 & 27\\
\bottomrule
\end{tabular}
\end{table}

\begin{table}[!t]
\caption[Experimental results for the regression (RMSE averaged for 20 runs).]{Experimental results for the regression (RMSE averaged for 20 runs). Best
results are marked in bold (lower is better). Offered method DV alone or in combination with others methods of regularization show best result for every dataset tested. Hypothesis about dependencies between complexity of dataset and best method was not proved.}
	\renewcommand{\arraystretch}{1.5}
	\fontsize{7pt}{7pt}\selectfont
	\begin{tabular}{p{1.3cm}| m{1.2cm}m{1.2cm}m{1.2cm}m{1.2cm}m{1.2cm}m{1.2cm}m{1.2cm}m{1.2cm}}
	\toprule
	\textbf{Method  } & \textbf{Air} &  \textbf{Boston} &   \textbf{Bike}  &
	\textbf{Energy} &\textbf{Sklearn} &   \textbf{House} & 
	\textbf{Scond} & \textbf{Crime} \\
	\textbf{Features} &   \textbf{10}&  \textbf{13} &   \textbf{13} &  \textbf{27} &
	\textbf{30} &    \textbf{81} & \textbf{81} & \textbf{100}\\

	\midrule
	No reg & 0.00880 &0.09496 & 0.03090 & 0.00506 &  0.06493 & 0.01363 & 0.08137 & 0.14596 \\
	\hline
	$L_2$ & 0.00620 &0.09122 & 0.02260 & 0.00431&  0.06088 & 0.01952  &0.07922 & 0.14270 \\
	\hline
	Dropout & 0.00744 &0.09121 & 0.01986 & 0.00455&  0.06435 & 0.01159 &  0.08100 & 0.14514 \\
\midrule
\midrule
	\textbf{DV gaus}&  0.00451 &  0.08958 &  \textbf{0.01566} & 0.00327&  0.06100 & 0.01133&  0.07930 & 0.14350 \\
	\hline
	\textbf{DV lapl} &0.00512 & 0.09093 & 0.02559 &-&  0.06106 & 0.01460  & -&-  \\
	\hline
	\textbf{DV anneal} &0.01207 & 0.09324 & 0.03013 & 0.00455&  0.06235 & \textbf{0.00967} &  \textbf{0.07867} & 0.14460  \\
	\hline
	\textbf{DE} & 0.00699& 0.08981 & 0.02525 & 0.00462 & 0.06106 & 0.01464 &  0.07953 & 0.14448 \\
\midrule
\midrule
	\textbf{DV+Drop} & 0.00453 & 0.090262 & 0.01960& \textbf{0.00153} &  0.06335 & 0.01504 &  0.07972 & 0.14389 \\
	\hline
	\textbf{DV+$L_2$} &0.00558 &\textbf {0.08728} & 0.02147 & 0.00313&  \textbf{0.06016} & 0.01142 &  0.07928 & \textbf{0.14215}   \\
	\hline
	\textbf{DV+DE} & \textbf{0.00293} &  0.08789 & 0.02327  & 0.00335 & 0.06496 & 0.01270 &  0.07937 & 0.14349  \\

    \bottomrule
\end{tabular}

    \label{reg_exper}
\end{table}

\textbf{Datasets.} We use eight datasets with different sizes and complexity for our evaluations: Boston House Prices (BHP) \cite{scikit-learn}, Bike Sharing (BS) \cite{bike}, Air Quality (AQ) \cite{DEVITO2008750}, Make-sklearn (MS) \cite{scikit-learn}, Housing Price (HP) \cite{de2011ames}, Superconductivty (SC) \cite{HAMIDIEH2018346}, Communities and Crime (CC) \cite{Dua:2019},  Appliances Energy Prediction (AEP) \cite{CANDANEDO201781}. The characteristics
of the datasets are summarized in Table \ref{regdata}. Also, Minmax scaling is used to standardize the features to present in the data in a fixed range.

\textbf{Architecture.} We implement our methods using neural network with two hidden layers and ReLU activation function. 

\textbf{Optimizer.} We use ADAM optimizer \cite{kingma2017adam} with learning rate of 0.001 in all experiments.

\textbf{Hyperparameters.} For each dataset a grid-search is used to find the best value for hyper-parameters:
\begin{enumerate}
     \item  $L_2$ penalty for $L_2$ regularization,
     \item  Drop rate (\%) for Dropout,
     \item  Disturb rate (\%) for Disturb Value,
     \item  Residual (e) for Disturb Error.
    
    \end{enumerate}

\textbf{Experimental Results.}  To tune hyper-parameters all datasets were split on training set (50\%) and testing set (50\%). For each hyper-parameter we fit our models on training set and then evaluate accuracy metric on testing set. For every run every dataset values were shuffled before split. Table \ref{reg_exper} shows the results of experiments, which proved that DV approach or combination of DV with DE, cosine-annealing, Dropout or $L_2$ outperforms baselines for all eight datasets. We were interested to know if the most suitable approach depends on data size and data complexity. To check this we form Table \ref{reg_exper} starting from a dataset with smaller complexity (smaller number of features) and ending with a dataset with bigger complexity. No dependencies were revealed during this analysis. 

We also analyze standard deviations for every method to be sure that offered approaches (DV and DE) are robust. There is no significant deviation between standard deviation for the model without regularization and models with regularization.

\section{Conclusion}
In this paper, we have extended one of the modern regularization methods, DisturbLabel (DL), by proposing an improved procedure of label disturbing for classification task and projecting the idea of useful ground truth disturbance to regression domain, which was not covered by the method of the baseline paper.

The problem of overfitting can be interpreted as memorizing the ground truth by a neural network. Our extension of the classification task is based on the hypothesis that confident predictions often come from this memorizing of the ground truth labels, that is why it makes more sense to penalize (in our case, randomly disturb) a share of them and do not perturb training procedure for the predictions that are unconfident in a natural way. We proposed to use the cosine similarity to measure the distance between vectors of ground truth and predicted class probabilities in classification tasks. We have tested this method on six datasets and two architectures and showed that DDL brings improvement compared to DL (alone or in combination with dropout).

We also have shown that the extension of DL method on the regression domain can improve the performance of the model and help to avoid overfitting. We presented two methods: (1) DisturbValue (DV) method which injects Gaussian noise to target values at random, and (2) DisturbError method (DE) which injects Gaussian noise to target values if the prediction is close to the target value, other words, the difference between prediction and target values is smaller than a small constant. 
 Our proposed DV method outperformed well-known baselines $L_2$ and dropout alone or in combination with other approaches (DE, $L_2$, dropout, cosine-annealing). The experiments were done for eight datasets with different sizes and complexity.



\end{document}